\documentclass[journal]{IEEEtran}
\usepackage{algorithm}
\usepackage[noend]{algpseudocode}
\newcommand\numberthis{\addtocounter{equation}{1}\tag{\theequation}}
\usepackage{amsmath}
\usepackage{mathtools}
\usepackage{amsmath,amsfonts,amssymb}
\usepackage{booktabs}
\usepackage{mathrsfs}
\usepackage{makecell}
\usepackage{multirow}
\usepackage{xcolor}
\usepackage[noadjust]{cite}
\DeclarePairedDelimiter\ceil{\lceil}{\rceil}
\DeclarePairedDelimiter{\round}\lfloor\rceil
\usepackage[switch,columnwise]{lineno}
\usepackage{amssymb}
\usepackage{pifont}
\newcommand{\xmark}{\ding{55}}%
\newcommand{\cmark}{\ding{51}}%

\ifCLASSINFOpdf
\else
\fi
\begin{document}

%
\title{ResOT: Resource-Efficient Oblique Trees \\for Neural Signal Classification}
%
%

\author{Bingzhao~Zhu, Masoud~Farivar, and Mahsa~Shoaran,~\IEEEmembership{Member,~IEEE}
\thanks{B. Zhu is with the School
of Applied and Engineering Physics, Cornell University, Ithaca,
NY, 14853 USA (e-mail: bz323@cornell.edu).}
\thanks{M. Farivar is with Google, Mountain View, CA, 94043 USA.}
\thanks{M. Shoaran is with the Center for Neuroprosthetics (CNP) and Institute of Electrical Engineering, EPFL, Switzerland, and School
of Electrical and Computer Engineering, Cornell University, Ithaca, NY, 14853 USA.}}


%
%

\markboth{\textit{\MakeLowercase{accepted by}} IEEE Transactions on Biomedical Circuits and Systems}
{Shell \MakeLowercase{\textit{et al.}}: Bare Demo of IEEEtran.cls for IEEE Journals}
%



\maketitle

\begin{abstract}
Classifiers that can be implemented on chip with minimal computational and memory resources are essential for edge computing in emerging applications such as medical and IoT devices. This paper introduces a machine learning model based on oblique decision trees  to enable resource-efficient classification on a neural implant. By integrating model compression with probabilistic routing and implementing cost-aware learning, our proposed model could significantly reduce the memory  and  hardware cost compared to state-of-the-art models, while maintaining the classification accuracy. \textcolor{black}{We trained the resource-efficient oblique tree with power-efficient regularization (ResOT-PE)} on three neural classification tasks to evaluate the performance, memory, and hardware requirements. On seizure detection task, we were able to reduce the model size by \textcolor{black}{3.4$\times$} and the feature extraction cost by \textcolor{black}{14.6$\times$ compared to the ensemble of boosted trees}, using the intracranial EEG from 10 epilepsy patients. In a second experiment, we tested the ResOT-PE model on tremor detection for Parkinson's disease, using the local field potentials from 12 patients implanted with a deep-brain stimulation (DBS) device. We achieved a comparable classification performance as the state-of-the-art boosted tree ensemble, while reducing the model size and feature extraction cost by \textcolor{black}{10.6$\times$} and \textcolor{black}{6.8$\times$}, respectively. We also tested on a 6-class finger movement detection task using ECoG recordings from 9 subjects, reducing the model size by \textcolor{black}{17.6$\times$} and feature computation cost by \textcolor{black}{5.1$\times$}. The proposed model  can enable a low-power and memory-efficient implementation of classifiers for real-time neurological disease detection and motor decoding.
\end{abstract}

\begin{IEEEkeywords}
Oblique trees, resource-efficient, feature extraction, machine learning, neurological disease detection.
\end{IEEEkeywords}

%
\IEEEpeerreviewmaketitle

\section{Introduction}
%
%
%
%
\IEEEPARstart{R}\noindent ecently, the use of machine learning (ML) techniques has been extended to emerging challenges in  neural data processing, such as early symptom detection  \textcolor{black}{\cite{yoo20128, shoaran2018energy, zhang2015low, page2014flexible,zhang2015seizure, yao2020improved}}, brain image classification \cite{nayak2016brain},  and motor decoding \cite{chen2015128, yao2019enhanced}. With the help of domain-specific biomarkers, ML models have been used to classify neurophysiological signals  with limited training sets \textcolor{black}{\cite{zhu2019migraine, shoaran2018energy, page2014flexible, zhang2015low, yoo20128, zhang2014seizure,zhang2015seizure, yao2019enhanced, yao2020improved, chen2015128}} ---typically recorded through invasive or non-invasive electrodes--- while outperforming other conventional methods. 
However, although modern machine learning tools have shown promise in neural signal classification, their deployment on high-channel-count neural interfaces remains a challenge, given the tight power budget and stringent area and memory constraints for such implants \cite{shaikh2019towards}. \textcolor{black}{The alternative approach that consists in transmitting the extracted features from neural channels for off-the-body classification \cite{verma2010micro,zhang2010implantable}, has several drawbacks. First, similar to raw data transmission, this approach suffers from security and privacy concerns due to transmitting patients’ private data to external servers for processing. Second, while power demands for telemetry may be relaxed due to lower-dimension feature transmission, the high loop latency could be problematic for real-time and closed-loop operation of  implantable devices (e.g., for closed-loop activation of a therapeutic or sensory stimulation in the brain). Making local predictions, on the other hand, could enable the device to work everywhere irrespective of connectivity to external units, and decisions could be made more quickly \cite{kumar2017resource}, with no need for raw data or feature streaming. }

\begin{figure}[t]
  \centering
  \includegraphics[width=1\columnwidth]{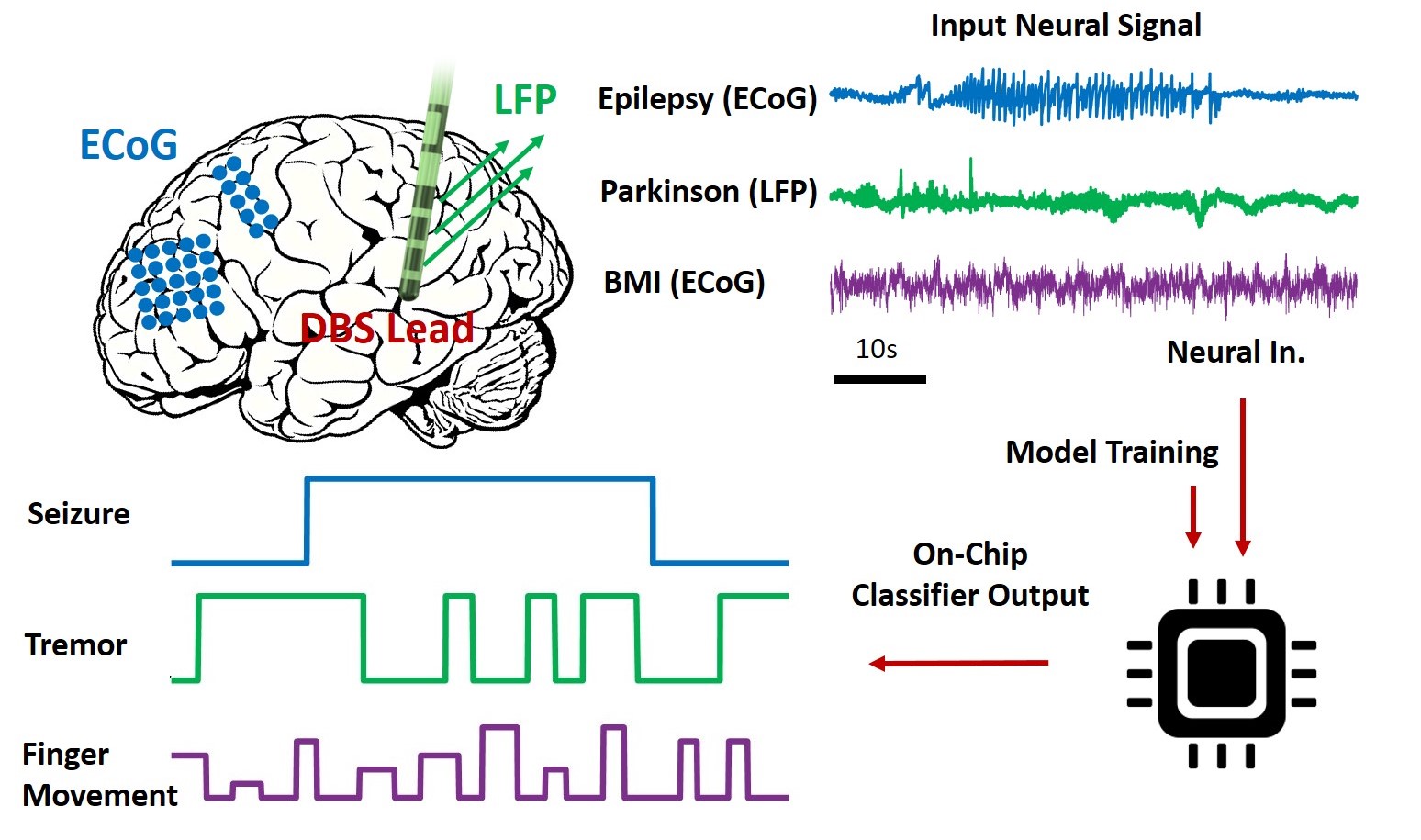}
  \caption{On-chip classification of neural signals for epileptic seizure detection, Parkinsonian tremor detection, or finger movement decoding. Neural signals are recorded from different regions of the brain. The machine learning model is trained offline and the parameters are stored on-chip. Decisions are made in real-time by the on-chip classifier, to predict a disease or classify a  movement. }\vspace{-5mm}
  \label{block}
\end{figure}

Among the widely-used ML algorithms for neural signal classification (e.g., logistic regression, support vector machines, k-nearest neighbours, neural networks, and decision trees), the latter is compatible with a lightweight, `on-demand' feature extraction framework, recently explored in \cite{shoaran2018energy}. A test sample travels through a single root-to-leaf path during inference, thus  visiting only a small proportion of the entire model \cite{shoaran2018energy, peter2017cost}. The lightweight feature extraction capability of decision tree (DT) is crucial, considering the large number of channels and predictive features in applications such as implantable seizure detection. In addition, according to recent studies \cite{kumar2017resource,carreira2018alternating}, the model size of DTs can be largely compressed to operate under extreme memory constraints. Given their lightweight inference and small model size, tree-based models can be  integrated on chip with state-of-the-art energy and area efficiency \cite{shoaran2018energy}, and are therefore favorable for neural and brain-machine interface applications. Furthermore, through techniques such as gradient boosting \cite{chen2016xgboost}, decision tree ensembles have achieved a high accuracy in classifying time-series neurophysiological data \textcolor{black}{\cite{kuhlmann2018epilepsyecosystem, shoaran2018energy, yao2020improved,yao2019enhanced,zhu2019migraine}}, typically outperforming the neural network-based models~\cite{taghavi2019hardware}.
We previously introduced a cost-efficient classification approach to further reduce the inference overhead of DTs, by adding the feature cost (e.g., power dissipation) as a regularization term to the objective function \cite{zhu2019cost}. In this cost-efficient learning scheme, each feature is associated with a hardware cost and the model is trained to prioritize the lower-cost features. 
This led to a reduction of  power dissipation by more than half for both seizure and tremor detection tasks, with only a marginal loss in performance (0.9\%) \cite{zhu2019cost}. 

However, building optimal binary DTs is essentially an NP-hard problem \cite{laurent1976constructing} and many approaches were recently proposed to optimize the tree structure \cite{norouzi2015efficient,hu2019optimal}. Unlike conventional axis-parallel trees that hold  deterministic decision functions,  probabilistic (i.e., soft) trees with oblique boundaries hold a probability decision leading  to the left or right child. 
Inspired by back-propagation neural networks, such probabilistic trees with stochastic routing are compatible with gradient-based optimization \cite{hehn2018end}. As a result, one may effectively employ various model compression techniques such as fixed-point quantization \cite{lin2016fixed}, weight pruning, and sharing \cite{han2015deep} that are widely used in  hardware implementation of deep neural networks (DNNs). 
Moreover, these soft DTs  still enable a lightweight inference, by following the most probable path along the tree. 

Here, we propose a framework based on soft oblique trees, by coupling neural networks with decision trees. Therefore, we can exploit the benefits of both models and compress an oblique tree (OT) with similar techniques as employed in DNN architectures. 
In addition, we extend our cost-efficient approach \cite{zhu2019cost} to soft oblique trees, as a promising alternative to conventional axis-aligned DTs. With these techniques, we demonstrate the performance of our single cost-aware oblique tree  on several neural classification problems including seizure, tremor, and finger movement detection tasks (Fig.~\ref{block}), and benchmark it against state-of-the-art models.

This paper is organized as follows. Section II describes the patient information, neural data, and classification tasks. Section III introduces the feature extraction process and probabilistic training for oblique trees. Several techniques to reduce the hardware cost including power-efficient learning, weight pruning, and sharing are  introduced in Section IV, followed by classification results in Section V. Discussions are presented in Section VI and Section VII concludes the paper.

\begin{table}[h]
\caption{Information on Patients and Neural Recordings}
\vspace{-5mm}
\begin{center}
\scalebox{0.81}{
\begin{tabular}{c|c|c|c}
\hline \hline
\textbf{Epilepsy}& \textbf{{\# of  Channels/}}& \textbf{{\# of}}& \textbf{{Recording}} \\
\textbf{iEEG Portal ID}& \textbf{{Sample Rate (Hz)}}& \textbf{{Seizures}}& \textbf{{Duration}} \\
\hline
I001\_P034\_D01 &47/5000 &16 &1d8h  \\
Study 004-2 &56/500  &3 &7d18h  \\
Study 022 &56/500  &7 &3d23h  \\
Study 024 &88/500  &19 &8d10h  \\
Study 026 &96/500  &22 &3d3h  \\
Study 029 &64/500  &3 &5d1h  \\
Study 030 &64/500  &8 &5d23h  \\
Study 033 &128/500  &17 &6d17h  \\
Study 037 &80/500  &8 &8d23h  \\
Study 038 &88/500  &10 &3d0h  \\
\hline \hline 

\textbf{Parkinson}& \textbf{\# of  Channels/}& \textbf{{Recording}}& \textbf{Duration (min)/} \\
\textbf{Recording Index}& \textbf{Sample Rate (Hz)}&  \textbf{{Side}}& \textbf{Tremor Prevalence (\%)} \\
\hline
1 &4/2048 &R  &5.7/81.0  \\
2 &4/2048 &L  &8.3/48.9  \\
3 &4/2048 &R  &6.6/40.9 \\
4 &4/2048 &L  &6.1/95.7  \\
5 &4/2048 &L  &4.9/45.3  \\
6 &4/2048 &L  &5.5/83.4  \\
7 &4/2048 &R  &9.6/52.3  \\
8 &4/2048 &L  &10.0/89.5  \\
9 &4/2048 &R  &10.0/94.0  \\
10 &4/2048 &L  &1.5/53.0  \\
11 &4/2048 &R  &4.5/79.4  \\
12 &4/2048 &R  &6.6/49.3  \\
13 &4/2048 &L  &4.8/96.5  \\
14 &4/2048 &L  &5.9/94.8  \\
15 &4/2048 &R  &5.0/78.0  \\
16 &4/2048 &L  &4.7/83.2  \\
\hline \hline

\textbf{Finger Movement}& \textbf{\# of  Channels/}& \textbf{Recording}& \textbf{Array} \\
\textbf{Subject Index}& \textbf{Sample Rate (Hz)}& \textbf{Side}& \textbf{Location} \\
\hline
1 &46/1000 &L &Fronto-Parietal  \\
2 &63/1000 &R &Fronto-Temporal   \\
3 &61/1000 &L &Fronto-Temporal-Parietal  \\
4 &58/1000 &L &Fronto-Temporal  \\
5 &64/1000 &L &Parietal-Temporal-Occipital  \\
6 &43/1000 &L &Fronto-Temporal  \\
7 &64/1000 &L &Fronto-Temporal-Parietal  \\
8 &38/1000 &R &Fronto-Parietal  \\
9 &47/1000 &L &Frontal   \\
\hline \hline 
\end{tabular}
}
\label{tab1}
\end{center}
\vspace{-7mm}
\end{table}

\vspace{-3mm}
\section{Neural Classification Tasks \& Data Description}
In this work, our focus is on algorithm development and hardware-algorithm co-optimization of resource-efficient oblique trees (ResOT) as a promising approach for  neural  signal  classification. To show the broad application and effectiveness of the proposed model, we evaluate this approach on three implantable neural applications described below, including epileptic seizure detection, Parkinsonian tremor detection, and finger movement classification, as depicted in the general block diagram of Fig.~\ref{block}.
\vspace{-3mm}
\subsection{Seizure Detection} Our first target application is seizure detection for medically refractory epilepsy, using continuous neural recordings from human subjects. Seizure detection is a binary supervised classification problem with the aim of classifying between  seizure and non-seizure states of a patient. We applied our model to the intracranial EEG (iEEG) recordings (shown as ECoG on Fig.~\ref{block}) from patients with epilepsy, publicly available on the iEEG portal~\cite{wagenaar2015collaborating}. \textcolor{black}{The dataset includes the iEEG recordings from 10 patients with at least 1 day of uninterrupted recording and 3 seizure events.} A total number of 113 seizure events were annotated by expert neurologists, as detailed in Table. \ref{tab1}.
The original recordings were segmented into 1-second  windows labeled as \textit{seizure} or \textit{non-seizure}.
\vspace{-3mm}
\subsection{Tremor Detection}
In a second study, we analyzed 16 local field potential (LFP) recordings from 12 patients with Parkinson's disease (PD) who were implanted with   4-channel deep-brain stimulation (DBS) lead(s), as described in \cite{yao2018resting, yao2020improved}. The statistics of this dataset are summarized in Table. \ref{tab1}. The LFP signals were recorded  from the subthalamic nucleus (STN) region at a 2048 Hz sampling rate. We labeled the LFP recordings as \textit{tremor} or \textit{non-tremor} based on the simultaneous acceleration measurements. Our ML model was  trained to differentiate between  \textit{tremor} and \textit{non-tremor} states, using 3 bipolar LFP channels.  The patients were recruited from the University of Oxford and gave informed consent to participate in the study that was approved by the local research ethics committee \cite{yao2020improved}.
\vspace{-5mm}
\subsection{Finger Movement Classification }
Our third study was focused on a finger movement classification task for brain-machine interface (BMI) application, using the electrocorticography (ECoG) data from 9 subjects sampled at 1kHz (Table. \ref{tab1}) \cite{miller2012human}. During experiments, subjects were asked to move one of their fingers for 2s, as instructed on a monitor. Overall, each subject performed 30 trials per finger. The finger movement was captured by a data-glove at 25 Hz. All patients participated in a purely voluntary manner after providing informed consent, under experimental protocols approved by the Institutional Review Board of the University of Washington (\#12193) \cite{miller2012human}. 
Unlike seizure and tremor detection tasks, the finger movement detection is a 6-class problem. The labels for this study were defined as \textit{thumb, index, middle, ring, and little finger movement} plus a \textit{rest state}.  
\vspace{-6mm}
\section{Model Description and Related Work}
The majority of current on-chip classifiers for neural signal analysis is based on support vector machines (SVMs). The EEG-based embedded seizure detectors in \cite{yoo20128,altaf201516,lee2013low} achieved energy efficiencies of 2.03, 1.85, and 273 $\mu$J/class, respectively, using SVM classifiers. 
\textcolor{black}{An incremental-precision algorithm  was proposed  to  reduce the energy  consumption for on-chip seizure detection  \cite{koteshwara2018incremental}, replacing the complex SVM  with logistic regression.}
Compared to SVM, DTs also offer a lightweight inference and can improve the energy efficiency for implantable applications, where resource constraints are more critical than EEG-based wearable systems. We recently improved the energy efficiency for on-chip seizure detection  to 41.2 nJ/class \cite{shoaran2018energy}, using an ensemble of eight gradient-boosted trees that required 1kB of memory to store model parameters.

While previous SoCs integrate a small number of trees  (e.g., $<$10 for seizure detection \cite{shoaran2018energy}, one for voice activity detection \cite{badami201590}), larger ensembles may be necessary for reliable detection of more complex symptoms (e.g., 30 trees for tremor detection  \cite{yao2020improved, yao2018resting}, 100 for finger movement classification \cite{yao2019enhanced}, and 105 for migraine ictal vs. interictal detection \cite{zhu2019migraine}). However, as shown in \cite{shoaran2018energy}, the power consumption and area of the classifier could linearly scale with the number of trees \cite{shoaran2018energy}. 

To address this challenge and reduce the number of correlated trees, we propose to employ oblique decision trees in our model, as depicted in Fig. \ref{neurtree}. 
This enables us to further improve the memory,  power efficiency, and scalability of our classifier. 
Oblique trees require fewer splits and learn through powerful nodes that employ more than one attribute to conduct a split, resulting in reduced number of trees. 
Such oblique nodes are effective in separating the highly-correlated features in our neural processing tasks and impose a modest  cost in hardware, as later discussed in this paper.
In addition, through gradient-based training as in neural networks, we are able to employ weight pruning and sharing techniques to compress the model and generate resource-efficient oblique trees. In this paper, we benchmark the ResOT approach against boosted tree ensembles \cite{shoaran2018energy, zhu2019cost}, as well as recent models based on sparse oblique trees, such as TAO \cite{carreira2018alternating} and Bonsai~\cite{kumar2017resource}. 

\vspace{-3mm}
\subsection{Neural Biomarker Extraction}
Unlike end-to-end learning approaches such as DNN, feature-engineered models rely on hand-crafted features and generally obtain a superior performance on small datasets \cite{zhu2019migraine,yao2020improved}. Here, we extract a set of predictive biomarkers for each task, followed by  ResOT  training and classification in a supervised manner.

Based on prior studies on the informative biomarkers of EEG/iEEG to predict seizures, we extracted the following features for our epilepsy task \cite{shoaran2018energy}: line-length (LLN), total power (Pow), variance (Var), and band power over delta ($\delta$: 1--4 Hz), theta ($\theta$: 4--8 Hz), alpha ($\alpha$: 8--13 Hz), beta ($\beta$: 13--30 Hz), low-gamma ($\gamma_1$: 30--50 Hz), gamma ($\gamma_2$: 50--80 Hz), high-gamma ($\gamma_3$: 80--150 Hz), ripple (R: 150--250 Hz) and fast ripple (FR: 250--600 Hz) bands, where fast ripples are only extracted from iEEG with 5 kHz sampling rate. 

For the second task, we used a set of predictive biomarkers of tremor in  LFP, based on our recent study on Parkinson's disease \cite{yao2020improved, yao2018resting}: the power of beta,  gamma, and high-frequency oscillation (HFO) in several sub-bands ($\beta_1$: 13--20 Hz, $\beta_2$: 20--30 Hz), ($\gamma_1$: 30--45 Hz, $\gamma_2$: 60--90 Hz, $\gamma_3$: 100--200 Hz), ($\text{HFO}_1$: 200--300 Hz, $\text{HFO}_2$: 300--400 Hz), the power ratio between low and high HFO ($\text{HFO}_R$), tremor power (TPow), and Hjorth parameters \cite{hjorth1970eeg}. The Hjorth activity represents the signal variance, Hjorth mobility indicates the mean frequency, and the rate of frequency changes is measured by Hjorth complexity  \cite{hjorth1970eeg}.

For finger movement classification task, we computed the power of ECoG over alpha ($\alpha$: 8--13 Hz), beta ($\beta$: 13--30~Hz), low-gamma ($\gamma_1$: 30--60 Hz), gamma ($\gamma_2$: 60--100 Hz) and high-gamma ($\gamma_3$, 100--200 Hz) bands, local motor potential (LMP) as the moving average of raw ECoG signal, and the Hjorth parameters (Act, Mob, Com)  \cite{yao2019enhanced}. A brief mathematical description of these features is given in Table.~\ref{tab2}. 

\begin{table}[t!]
\caption{Computed Features \&  Normalized Power Cost}
\vspace{-4mm}
\begin{center}
\vspace{-0mm}
\scalebox{0.85}{
\begin{tabular}{c|c|c}
\hline \hline
\textbf{Epilepsy} & \textbf{Description} &  \textbf{Power}\\
\hline
Line-Length (LLN) & \hspace{-2mm} $\frac{1}{d} \sum_{d}|x[n]-x[n-1]|$, $d=$window size \hspace{-2mm} &   1 \\
Power (Pow) & $\frac{1}{d} \sum_{d}x[n]^2$ &   1.87\\
Variance (Var) &$\frac{1}{d} \sum_{d}(x[n]-\mu)^{2},$ $ \mu=\frac{1}{d} \sum_{d}x[n]$& 2.93\\
Delta ($\delta$) & Band power in 1--4 Hz & 34.07 \\
Theta ($\theta$) & Band power in 4--8 Hz &  34.07 \\
Alpha ($\alpha$) & Band power in 8--13 Hz &  34.07 \\
Beta ($\beta$) & Band power in 13--30 Hz & 34.07 \\
Low-Gamma ($\gamma_1$)  &  Band power in 30--50 Hz & 34.07 \\
Gamma ($\gamma_2$) & Band power in 50--80 Hz &  34.07 \\
High-Gamma ($\gamma_3$) &  Band power in 80--150 Hz & 34.07 \\
Ripple (R) &  Band power in 150--250 Hz &  34.07 \\
Fast Ripple (FR) &  Band power in 250--600 Hz  (at 5 kHz)  & 34.07 \\

\hline \hline
\textbf{Parkinson} & \textbf{Description} &  \textbf{Power}\\
\hline
Low-Beta ($\beta_1$)  &Band power in 13--20 Hz &  34.07 \\
High-Beta ($\beta_2$) & Band power in 20--30 Hz &  34.07 \\
Low-Gamma ($\gamma_1$) & Band power in 30--45 Hz &  34.07 \\
Gamma ($\gamma_2$) & Band power in 60--90 Hz &  34.07 \\
High-Gamma ($\gamma_3$) & Band power in 100--200 Hz &  34.07 \\
Low-HFO ($\text{HFO}_1$) & Band power in 200--300 Hz &  34.07 \\
High-HFO ($\text{HFO}_2$) & Band power in 300--400 Hz & 34.07 \\
HFO Ratio ($\text{HFO}_R$) & Low-HFO to High-HFO ratio & 68.15 \\
Tremor Power (TPow) & Band power in 3--7 Hz & 34.07 \\
Hjorth Activity (Act) &  $\frac{1}{d} \sum_{d}(x[n]-\mu)^{2},$ $ \mu=\frac{1}{d} \sum_{d}x[n]$ & 2.93\\
Hjorth Mobility (Mob) & $\sqrt{\frac{\text{Var}(x[n]-x[n-1])}{\text{Var}(x[n])}} $ &  6.26\\
Hjorth Complexity (Com) & ${\frac{\text{Mob}(x[n]-x[n-1])}{\text{Mob}(x[n])}} $ & 9.62  \\

\hline \hline
\textbf{Finger Movement} & \textbf{Description} &  \textbf{Power}\\
\hline
Alpha ($\alpha$) & Band power in 8--13 Hz &  34.07 \\
Beta ($\beta$)  &Band power in 13--30 Hz &  34.07 \\
Low-Gamma ($\gamma_1$) & Band power in 30--60 Hz &  34.07 \\
Gamma ($\gamma_2$) & Band power in 60--100 Hz &  34.07 \\
High-Gamma ($\gamma_3$) & Band power in 100--200 Hz &  34.07 \\
Local Motor Potential (LMP) & $\frac{1}{d} \sum_{d}x[n]$& 0.50 \\
Hjorth Activity (Act) &  $\frac{1}{d} \sum_{d}(x[n]-\mu)^{2},$ $ \mu=\frac{1}{d} \sum_{d}x[n]$ & 2.93\\
Hjorth Mobility (Mob) & $\sqrt{\frac{\text{Var}(x[n]-x[n-1])}{\text{Var}(x[n])}} $ &  6.26\\
Hjorth Complexity (Com) & ${\frac{\text{Mob}(x[n]-x[n-1])}{\text{Mob}(x[n])}} $ & 9.62  \\
\hline \hline
\end{tabular}
}
\label{tab2}
\end{center}
\vspace{-4mm}
\end{table}

\vspace{-3mm}
\subsection{Feature Cost Estimation}
\label{hce}
To implement a cost-aware ML model (as described in Section~IV), we first analyzed the hardware cost associated with different features, by simulating the power consumption for each individual feature. A standard digital implementation with 1.2V supply was used for circuit simulations. To extract band power features, FIR filters with 30 taps, 8-bit coefficients, and a parallel architecture were implemented. We have previously shown that 30 taps is a reasonable choice for FIR filters, considering the trade-off between hardware complexity and classification accuracy  \cite{zhu2019hardware}.
The design was synthesized in a 65nm TSMC LP process and the power consumption (post place and route) is reported in Table.~\ref{tab2}, after normalizing to the power of line-length, as the lowest complexity feature for the first task. 
\vspace{-3mm}
\subsection{Oblique Trees with Probabilistic Splits}
\label{Probabilistic}
DTs are among the most powerful ML models that are widely used in practice. A tree is composed of basic computational units, called internal nodes and leaf nodes. Characterized by the hierarchical scheme of the nodes, DTs fit into complex nonlinear distributions. We previously implemented a gradient boosting ensemble of eight axis-aligned trees to detect epileptic seizures \cite{shoaran2018energy}. The chip was implemented in a 65nm TSMC process and achieved state-of-the-art performance in terms of energy-area-latency product \cite{shoaran2018energy}. Here, we consider a classification task with an input space of $ \mathcal{X} \subset \mathbb{R}^{D}$ and output space of $\mathcal{Y} = \{1,...,K\}$. Alternatively, the goal of this work is to learn a probabilistic  tree model that can map from the feature space ($\mathcal{X}$) to the label space ($\mathcal{Y}$). 

As opposed to axis-aligned decision trees, oblique trees use multiple features to make splits. \textcolor{black}{As shown in Fig. \ref{neurtree}, two-layer neural networks are used as split function to combine multiple features in each internal node.} As a result, the hyperplane is oblique rather than axis-aligned and can better fit to data with correlated features \cite{menze2011oblique}. 
Training oblique trees is not a trivial task, as the weights are not differentiable. In addition, without compressing the tree structure, oblique trees may grow overly complex and use many features to make splits, increasing both the model size and node complexity.
To tackle these issues, we introduce the oblique trees with probabilistic splits. Rather than deterministically routing a decision tree, we send a sample to the left or right subtree based on a probability value. \textcolor{black}{With this probabilistic routing, we can derive the objective function and train oblique trees with gradient-based optimization algorithms, and use various compression techniques applied to neural networks.}
\subsubsection{Data} We consider a dataset with $N$ instances $\{(\boldsymbol{x_1},y_1),...,(\boldsymbol{x_N},y_N)\} \subset (\mathcal{X},\mathcal{Y})$ where $\boldsymbol{x_n}$ is a feature vector of length $D$ and $y_n$ indicates the corresponding label.
\subsubsection{Internal node} 
\begin{figure}[t]
  \centering
  \includegraphics[width=1\columnwidth]{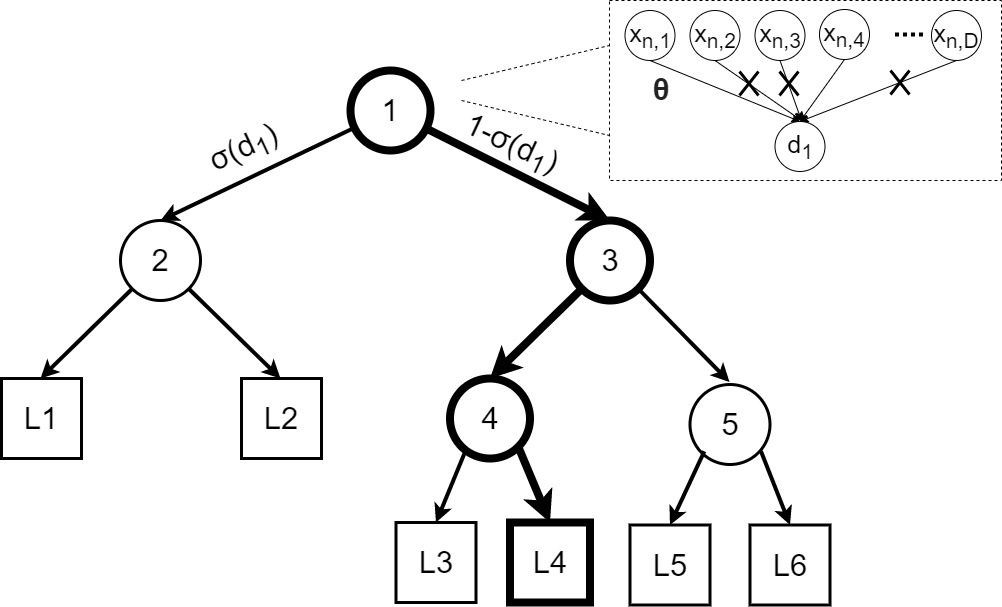}
  \vspace{-5mm}
  \caption{Proposed oblique tree, trained with soft decisions.  In the inference phase, the test samples follow the most probable path along the tree. Inside internal nodes, the decision functions can be represented by a two-layer neural network, for which we use weight pruning and sharing techniques to create sparse connections.}
  \label{neurtree}\vspace{-1mm}
\end{figure}

In a binary decision tree, each internal node $i \subset \{1,...,I\}$ has two child nodes. Generally, the internal nodes compute a binary function leading to the left or right child. On the other hand, in a probabilistic tree, the internal nodes make a soft decision which generates the probability of that split going to the left or right. Since the tree is oblique, we model the decision function as:
\begin{gather}
     \sigma(d_i(\boldsymbol{x_n})) =\frac{1}{1+e^{-\boldsymbol{{x}_n}^{\!\!\!\top}  \boldsymbol{\theta_i}}}
\end{gather}
where $\boldsymbol{\theta_i}$ is the weight vector for the $i$-th internal node, $\sigma(x)$ denotes the \textit{sigmoid} function, $d_i(\boldsymbol{x_n})$ is the output of internal node $i$, and $ \sigma(d_i(\boldsymbol{x_n}))$ indicates the probability of  leading the   feature vector $\boldsymbol{x_n}$ to the left child at node $i$, Fig.~\ref{neurtree}. 

\subsubsection{Leaf node} Leaf nodes are the terminal nodes of a decision tree. Each leaf node can be reached through a unique path which follows a set of decisions made by the internal nodes. Here, we use $\mathcal{R}_{\ell} \subset\{1, \ldots, I\}$ to represent the internal nodes which contain leaf node $l$ in the right subtree, and $\mathcal{L}_{\ell} \subset\{1, \ldots, I\}$ to denote the internal nodes which contain leaf node $l$ in the left subtree. Hence, the probability of the feature vector $\boldsymbol{x_n}$ reaching leaf node $l$ can be expressed as:
\begin{gather}
\label{equation2}
p\left(l | \boldsymbol{x_n} ; \boldsymbol{\theta}\right)=\prod_{i \in \mathcal{L}_{\ell}} \sigma\left(d_i(\boldsymbol{x_n})\right) \prod_{i \in \mathcal{R}_{\ell}}\left(1-\sigma\left(d_i(\boldsymbol{x_n})\right)\right).
\end{gather}


\vspace{-3mm}
\subsection{Learning Procedure}
Let us consider a supervised learning problem with $N$ pairs of samples $(\boldsymbol{x_n},y_n)$. Our goal is to maximize the empirical log-likelihood of the training data:
\begin{gather}
    \max _{\boldsymbol{\theta},\boldsymbol{\omega}} \sum_{n=1}^{N} \log p(y_n | \boldsymbol{x_n};\boldsymbol{\theta},\boldsymbol{\omega}).
\end{gather}

\noindent where ${\omega_{l,k}}$ indicates the probability of leaf $l$ having the class label $k$. The ${\omega_{l,k}}$ is normalized so that the sum of probabilities in a leaf is 1, i.e., $\sum_{k=1}^{K}{\omega_{l,k}}=1$. In the probabilistic routing scheme, $p(y_n | \boldsymbol{x_n};\boldsymbol{\theta},\boldsymbol{\omega})=\sum_{l=1}^{L} p(l|\boldsymbol{x_n};\boldsymbol{\theta}) \omega_{l,y_n}$. Combining the empirical loss with the regularization term $\lambda \Omega$ (explained in the following Section), the optimization objective of our model can be expressed as follows:
\begin{gather} \label{eq5}
O(\boldsymbol{\theta},\boldsymbol{\omega};\mathcal{X},\mathcal{Y})=\sum_{n=1}^{N} \log \sum_{l=1}^{L} p(l|\boldsymbol{x_n};\boldsymbol{\theta}) \omega_{l,y_n} + \lambda \Omega.
\end{gather}
The training process for probabilistic trees is  to maximize the  above objective function. A detailed discussion on estimating the optimal parameters ($\boldsymbol{\theta^*}$, $\boldsymbol{\omega^*}$) can be found in \cite{hehn2018end}. \textcolor{black}{Intuitively, by maximizing Eq. \ref{eq5}, the samples ($\boldsymbol{x_n}, y_n$) are encouraged to reach a leaf node $l$ where the probability of class label $\omega_{l,y_n}$ is maximized.} Here, we use the gradient-based Adam optimizer \cite{kingma2014adam} for learning soft oblique trees. The maximization of  $O(\boldsymbol{\theta},\boldsymbol{\omega};\mathcal{X},\mathcal{Y})$ is based on  subsets of training samples for which $\boldsymbol{\theta}$ and $\boldsymbol{\omega}$ are updated with mini-batches until $O(\boldsymbol{\theta},\boldsymbol{\omega};\mathcal{X},\mathcal{Y})$ converges.
\section{Weight Pruning, Sharing \& \\ Cost-Aware, Single-Path Inference}
\label{Prune}
Weight pruning and sharing are model compression techniques that were initially introduced \textcolor{black}{ to reduce the model size for DNNs \cite{lecun1990optimal,han2015deep}, while a number of recent studies have focused on hardware-efficient implementation of compression techniques  \cite{deng2018permdnn}.}
Considering that in the training phase of oblique trees we apply a gradient-based algorithm,  each internal node can be viewed as a two-layer fully connected network. Therefore, one may compress the oblique trees through similar techniques as those used in neural network compression. Moreover, trees are based on a hierarchical structure which is favored for lightweight inference, as predictions can be made following a single root-to-leaf path without visiting the complete model. By employing these techniques, we can significantly reduce the inference complexity of oblique trees and enable their resource-efficient integration on chip.
\vspace{-3mm}
\subsection{Regularization}
In a general neural network pruning scheme, small weights are considered trivial and set to zero. Regularization essentially penalizes the weights and encourages the values to be small. We explore two types of regularizations in this work: 1. The conventional $\ell_{2}$ regularization; 2. A new power-efficient regularization that attempts to minimize the feature computation cost (i.e., power consumption) during inference~\cite{zhu2019cost}.

\subsubsection{$\ell_{2}$ Regularization}
While $\ell_{1}$ regularization can generate sparse representations and has been used to optimize OTs \cite{carreira2018alternating}, it has been shown that $\ell_{2}$ regularization performs better for weight pruning \cite{han2015deep}. Inspired by the efforts on DNN compression, here we combine the $\ell_{2}$ regularization and weight pruning to generate sparse oblique trees. In the following, $\boldsymbol{\theta_{i}}$  represents the weight vector at the $i$-th internal node and the $\ell_{2}$ regularization term is expressed as:
\begin{gather}
\label{eq14}
    \Omega_{\ell_{2}}  = \sum_{i=1}^{I} |\boldsymbol{\theta_{i}}|^2.
\end{gather}

\subsubsection{Power-Efficient Regularization}
In order to design a cost-aware classifier \cite{peter2017cost}, we propose to include a regularization term that incorporates the hardware cost for feature extraction. In our previous work \cite{zhu2019cost}, we proposed a cost-aware model based on gradient-boosted tree ensemble to reduce the power consumption, while maintaining the classification accuracy. Here, we extend the cost-efficient study to oblique decision trees. In the following, $\boldsymbol{\beta}$ represents the feature cost vector of length $D$ and $p_{n,i}$ is the probability of instance $n$ going through internal node $i$. Therefore, the feature extraction cost for instance $n$ at node $i$ can be written as: 
\begin{gather}
    \Omega_{n,i}=p_{n,i} \boldsymbol{\beta}^\top  \| \boldsymbol{\theta_{i}}\|_{0}.
\end{gather}
\noindent where $\|\theta_{i}\|_{0}$ is the $\ell_{0}$ normalization indicating which features are used to make decisions. However, the $\ell_{0}$ normalization is not differentiable. Thus, it is not compatible with gradient-based training. Following an approach similar to \cite{xu2012greedy}, we instead approximate it with $\ell_{1}$ normalization. The new $\Omega_{n,i}$ is expressed as:
\begin{gather}
    \Omega_{n,i} \approx  p_{n,i} \sum_{j=1}^{D} \beta_j |\theta_{i, j}|
\end{gather}
\noindent where $\beta_j$ and $\theta_{i, j}$ indicate the $j$-th entry of the vectors $\boldsymbol{\beta}$ and $\boldsymbol{\theta_{i}}$, respectively. Here, $\Omega_{n,i}$ denotes the expected feature extraction cost for a test sample $n$ reaching the internal node~$i$. By evaluating the cost of  mini-batch on the entire tree, the power-efficient regularization term can be expressed as:
\begin{align*}
    \label{eq16}
        \Omega_{power-efficient} &= \frac{1}{N} \sum_{n=1}^{N} \sum_{i=1}^{I} \Omega_{n,i}\\
        &\approx \frac{1}{N} \sum_{n=1}^{N}  \sum_{i=1}^{I} p_{n,i} \sum_{j=1}^{D} \beta_j |\theta_{i, j}| \numberthis
\end{align*}
\noindent where $N$ and $I$ represent the number of training samples and internal nodes, respectively. This equation essentially introduces a power-dependent regularization term for the objective function in Eq.~\ref{eq5}, by estimating the power consumption for feature extraction based on the probability of visiting each node during training. The feature costs are averaged over samples to take into account the frequency of visiting a node during training (e.g., root nodes are visited more often than those in the deeper layers of a tree). 


\begin{figure}[t]
  \centering
  \includegraphics[width=0.99\columnwidth]{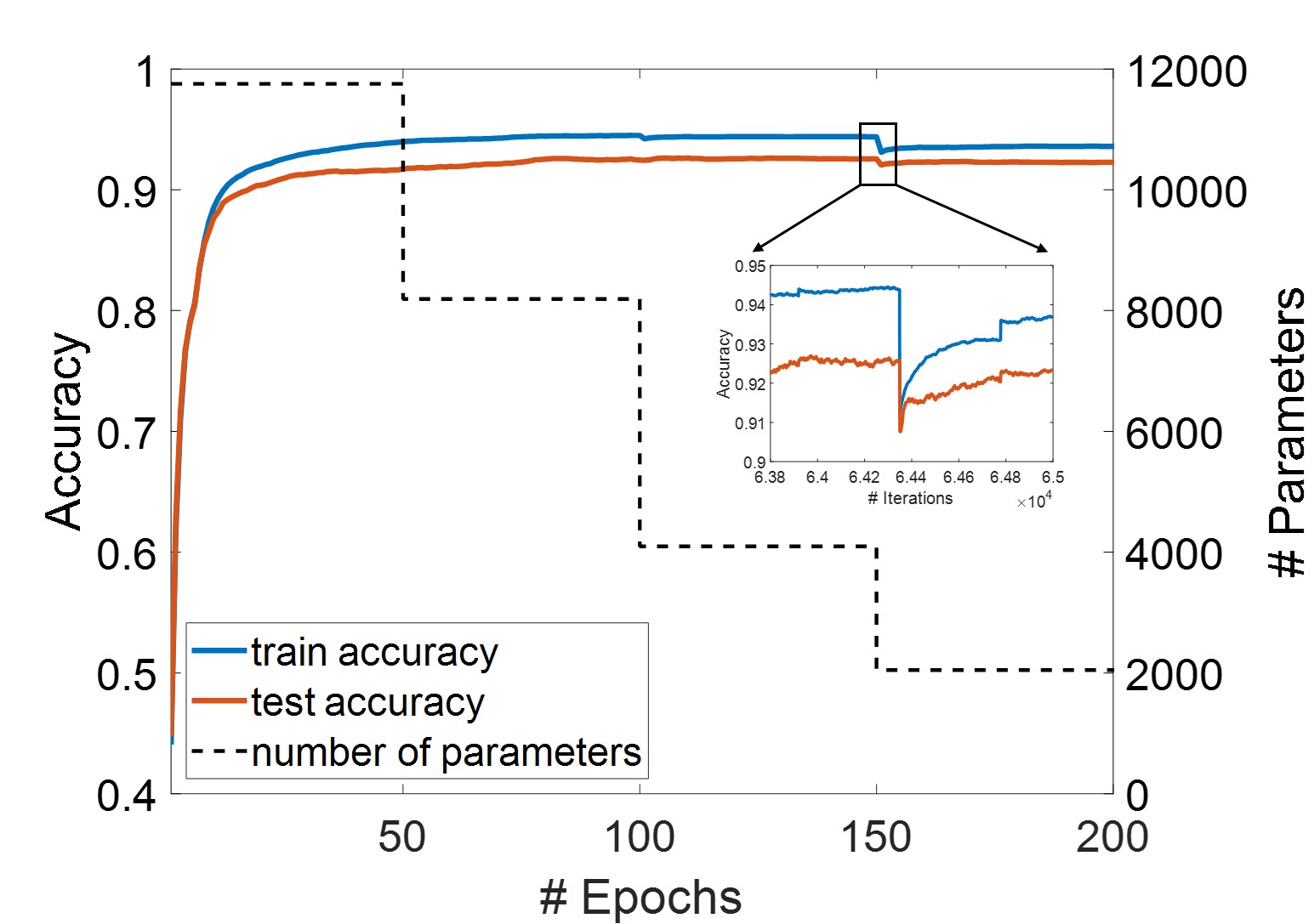}
  \caption{The prune-retrain process to generate sparse oblique trees. Weight pruning happens every 50 epochs. Testing on the MNIST dataset, the number of parameters is reduced by a factor of 5.7, using weight pruning.}
  \label{pruning}
\end{figure}

\vspace{-2mm}
\subsection{Weight Pruning}
The number of weights associated with each internal node equals the number of attributes. Therefore, the entire weight matrix $\boldsymbol{\theta}$ has a size of $ I \times D $. Each internal node contains a decision function which can be expressed as a fully connected network with only input and output layers. Here, we follow a train-prune-retrain process to generate an oblique tree with sparse connections in its internal nodes. In the pruning phase, small weights below a threshold are set to zero. Then, we retrain the tree to optimize the remaining weights. We iteratively repeat this prune-retrain process until the final performance converges, as shown in Fig. \ref{pruning}. 

Following weight pruning, we store the sparse weight matrix including the values and relative indices of the survived weights. The sparse weight matrix is stored in a column-first order and delta encoding is used to index the relative position of non-zero weights \cite{han2015deep}. To benchmark our approach against state-of-the-art oblique tree-based models (see Section V), we first tested the oblique tree with pruning  on the MNIST dataset. As depicted in Fig.~\ref{pruning}, we were able to reduce the parameter count from 11.8k to 2.0k with $<$0.1\% loss in accuracy.

\begin{figure}[t]
  \centering
  \includegraphics[width=1\columnwidth]{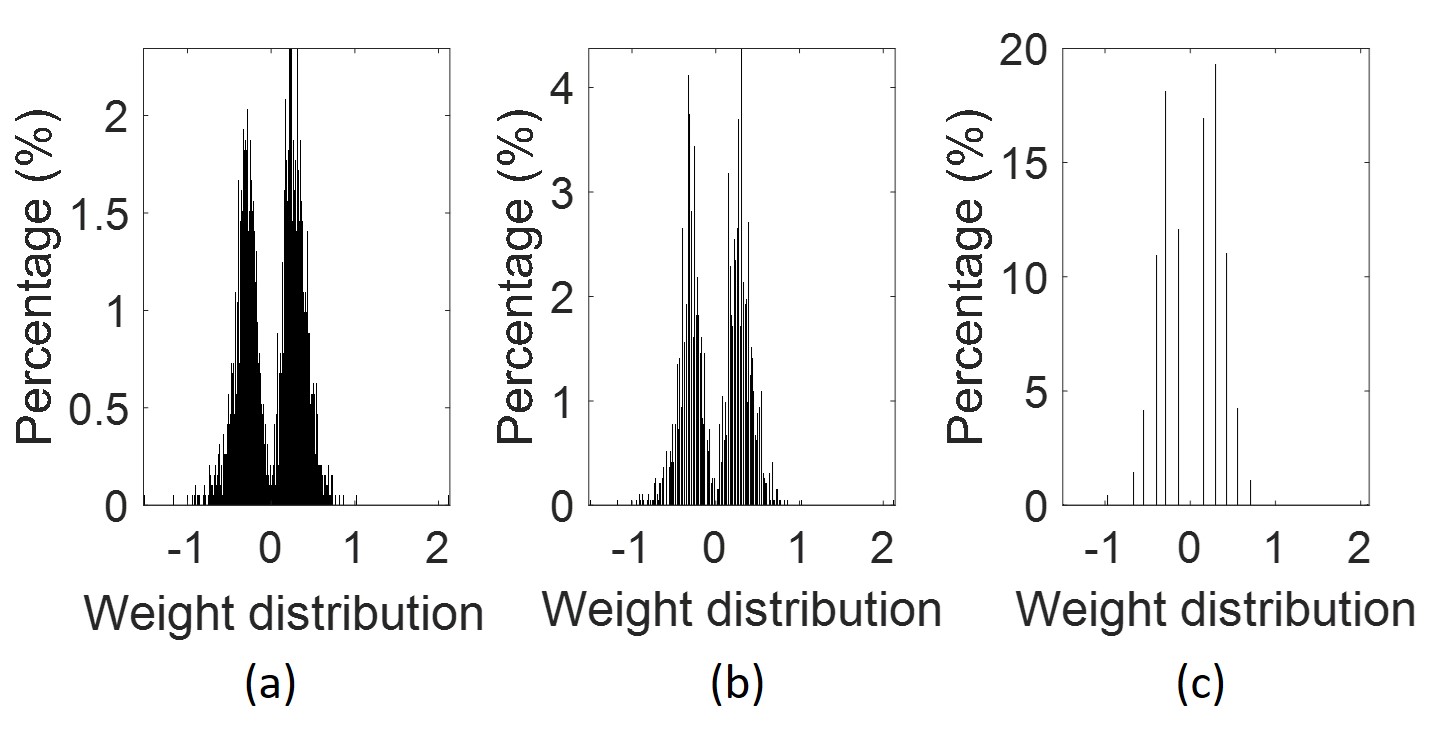}
  \vspace{-6mm}
  \caption{
  Weight distribution of an oblique decision tree: (a) without weight sharing, (b) after 8-bit weight sharing, and (c) after 4-bit weight sharing.}
  \label{quantization}
\end{figure}

\begin{algorithm}[b]
\caption{\textcolor{black}{Learning Resource-Efficient Oblique Trees}}\label{euclid}
\begin{algorithmic}
\Require{$T$: training set, $\boldsymbol{\beta}$: feature cost}
\State $\boldsymbol{\theta},\boldsymbol{\omega} \gets \text{pretrained by maximizing } Eq.(4)$
\For {$i \in \{1, \ldots, { nPrune}\}$} 
    \State $\boldsymbol{\theta} \gets WeightPruning(\boldsymbol{\theta})$
    \For {$k \in \{1, \ldots,  { nEpochs}\}$} 
        \State $\text{update } \boldsymbol{\theta},\boldsymbol{\omega} \text{ by maximizing } Eq.(4)$
    \EndFor
\EndFor
\State $\boldsymbol{\theta} \gets WeightSharing(\boldsymbol{\theta})$
\end{algorithmic} 
\end{algorithm}

\vspace{-2mm}
\subsection{Weight Sharing}
The weight sharing process implemented here is similar to the weight quantization approach proposed in \cite{han2015deep}. We first determine the range of the original weights. The weights are then uniformly separated into $k$ clusters across the entire range and the shared weights are initialized by the average weight of each cluster. Thus, all weights abandon their original values and take the value of the shared weight. Following this step, the possible weights are within $k$ shared values and  only  $ \ceil*{log_2 k}$ bits are required to index the weights, which are stored in memory as floating point numbers ($k \times 32$ bits required for shared weights). Here, $\ceil{x}$ indicates  rounding up $x$ to the nearest integer. Lastly, we fine-tune the shared weights through gradient-based optimization and correct the potential bias induced by direct weight sharing. Figure \ref{quantization} shows the weight distribution before and after weight sharing. 

\begin{figure}[t!]
    \centering
  \includegraphics[width=0.8\columnwidth]{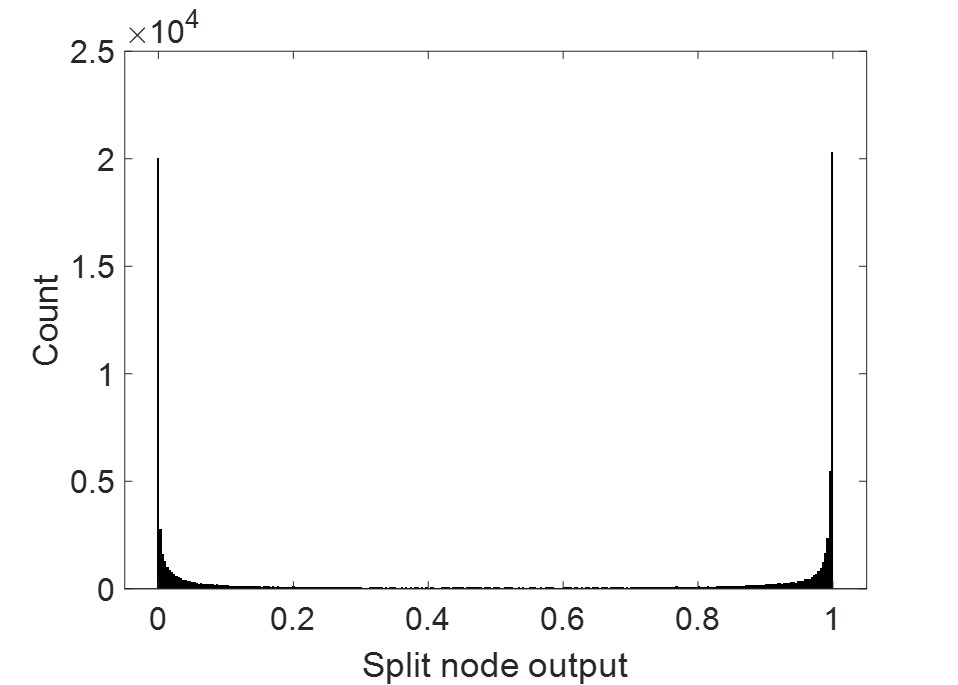} 
  \caption{Histogram of split node outputs. The x-axis indicates the probability of leading a sample to the left sub-tree. Decisions at the split nodes are mostly certain (0 or 1). The experiment is conducted on the MNIST dataset.}
  \label{re}
  \vspace{-4mm}
\end{figure}

\vspace{-5mm}
\subsection{Single-Path Inference}
In the common inference scheme for probabilistic trees, a test sample can travel through multiple paths, while the probability of that sample reaching each leaf node is calculated. The final prediction is made by averaging the leaf values based on their probabilities. This is referred to as a multi-path approach as shown in Eq. \ref{equation2}. 


Alternatively, we apply a single-path approach to enable lightweight inference. In the proposed scheme, the test samples choose the most probable path at each internal node, as shown in Fig. \ref{neurtree}. This enables us to make predictions by only visiting a small portion of the model and extracting fewer features. The single-path routing scheme is summarized in Eq.~\ref{eq15}, where $\round{p}$ denotes rounding the probability $p$ to either $0$ or $1$.
\begin{gather} \label{eq15}
\small
    p_{\text{single-path}}(l | \boldsymbol{x_n} ; \boldsymbol{\theta})=\prod_{i \in \mathcal{L}_{\ell}} \round{\sigma\left(d_{i}(\boldsymbol{x_n})\right)} \prod_{i \in \mathcal{R}_{\ell}}\round{1-\sigma\left(d_{i}(\boldsymbol{x_n})\right)}.
\end{gather}
However, the single-path inference is an approximation method and may result in information loss. To carefully analyze this, the histogram of all split node outputs for the proposed soft decision tree is plotted in Fig. \ref{re}. This distribution shows that samples are routed to the left or right sub-trees with very low uncertainty. In other words, most available paths will never be visited. Therefore, only the most probable path needs to be evaluated at the test time. Similar results were reported in \cite{tanno2018adaptive,hehn2018end} and our experimental results support this single-path inference scheme.

\textcolor{black}{The algorithmic pseudocode to train ResOT is presented as Algorithm.~\ref{euclid}. We performed the train-prune-retrain process for $nPrune$ rounds, and updated the parameters for $nEpochs$ epochs using gradient-based learning. By employing model compression techniques and gradient-based maximization of Eq. \ref{eq5}, the algorithm returns a sparse weight matrix $\boldsymbol{\theta}$ for an oblique decision tree.}

\begin{figure}
  \centering
  \includegraphics[width=0.82\columnwidth]{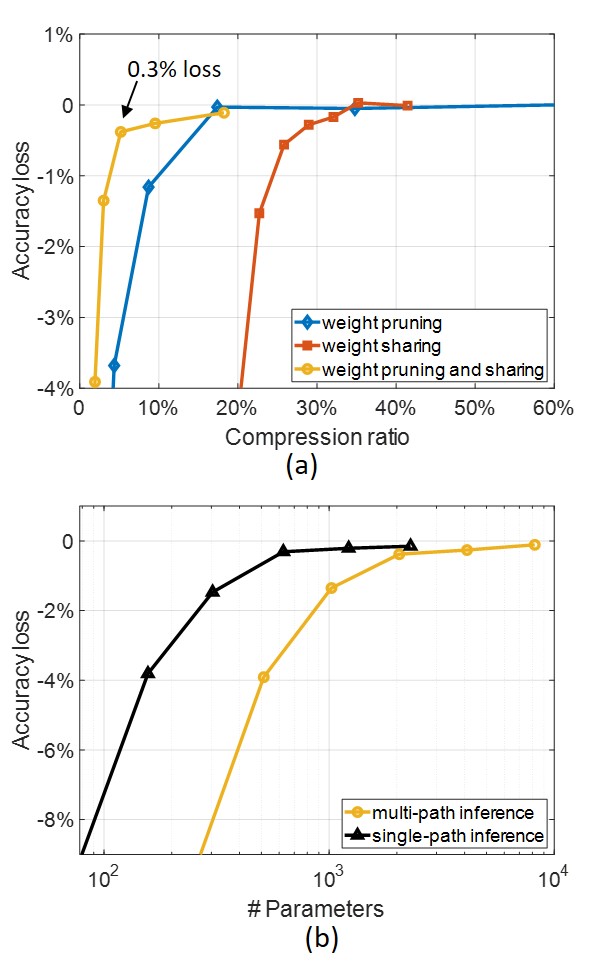}
  \caption{(a) The compression ratio of an oblique tree with weight pruning and sharing. We are able to compress the model by $20\times$ with a marginal accuracy loss (0.3\%); (b) Comparison between the single- and multi-path inference schemes in terms of average number of parameters used during inference. The single-path inference requires fewer parameters and causes no performance loss. The experiment is conducted on the MNIST dataset.}\vspace{-4mm}
  \label{comparison}
  \vspace{-4mm}
\end{figure}

 \vspace{-1mm}
\section{Results}
In this paper, we propose a framework to generate resource-efficient oblique trees (ResOT) with model compression, power-efficient learning \cite{zhu2019cost}, and single-path inference. As a result, our approach offers a small model size and lightweight inference, both requisite for low-power brain implants. In this Section, we implemented the ResOT model for several tasks, including a toy digit recognition task \cite{lecun1998gradient} on MNIST dataset and three neural signal classification  problems. The purpose of first task is to benchmark our model against state-of-the-art algorithms that are all tested on the MNIST dataset, in terms of accuracy and model compression. 

Among these tasks, seizure and tremor detection require binary labels, whereas digit recognition and finger movement classification involve multiple classes in their label space. A single ResOT was built for both binary and multi-class  tasks. In addition, we compared the ResOT approach with other DT-based models in terms of classification performance, model size, and power consumption, particularly for the three neural classification tasks that require a low-power implementation.

\begin{figure}[b]
  \centering
  \includegraphics[width=0.99\columnwidth]{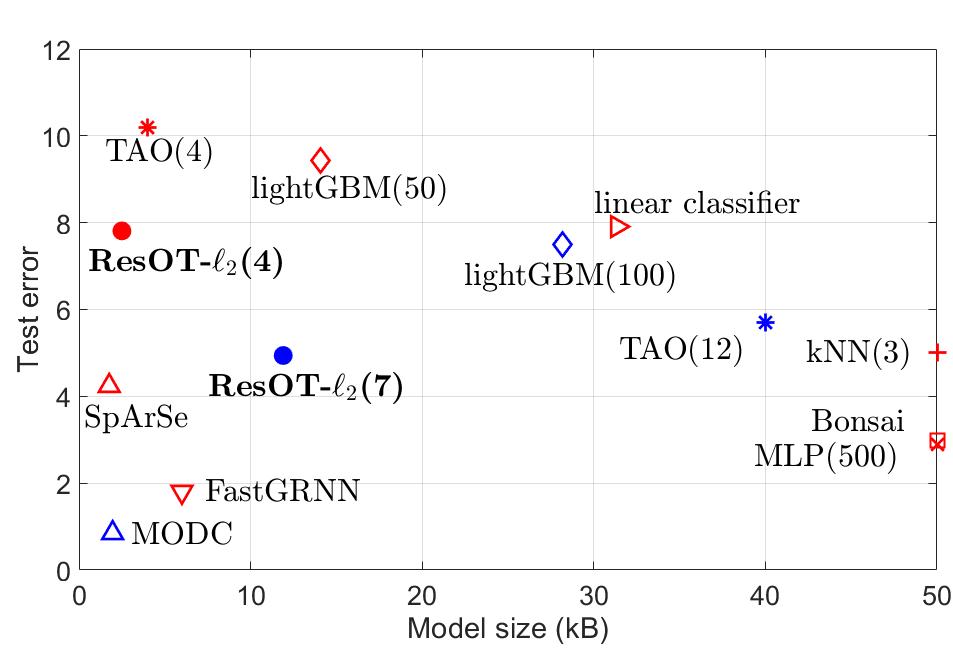}
  \caption{The performance of ResOT on the MNIST dataset compared to other algorithms based on sparse oblique trees (TAO \cite{carreira2018alternating} and Bonsai \cite{kumar2017resource}). Other machine learning models such as \textcolor{black}{FastGRNN \cite{kusupati2018fastgrnn}, SpArSe \cite{fedorov2019sparse}, MODC \cite{gural2019memory},} lightGBM \cite{ke2017lightgbm}, logistic regression (linear model), k-nearest neighbors (kNN) and multilayer perceptron (MLP) were also included. \textcolor{black}{Various model types are shown with different marker styles, while different colors represent model configurations.} Values outside the range are plotted on the boundary.}
  \label{mnist_compare}
\end{figure}

\vspace{-2mm}
\subsection{Performance on MNIST Dataset}
We first tested the oblique tree compression described in Section~IV on a toy dataset, MNIST \cite{lecun1998gradient}, for a 10-class digit recognition task. We used the $\ell_2$ regularization (Eq. \ref{eq14}) for weight pruning, since the feature extraction cost was not a concern for this task. \textcolor{black}{Testing on a computer with a 6-core i7-8700K CPU, 178.4 seconds were required to build a ResOT of depth 4  (pruning rounds: 4, batch size: 128, total epochs: 400).}
In order to reduce the model size, the tree was pruned to 2048 parameters, indexed by 4 bits following weight sharing. 
Compared with the original OT model, weight pruning and sharing  significantly reduced the model size. As illustrated in Fig.~\ref{comparison}(a), weight pruning and sharing could reduce the model size by $20\times$ with only a marginal accuracy loss (0.3\%). 
Moreover, although ResOT is trained with a probabilistic routing scheme, it is still compatible with lightweight single-path inference. In Fig.~\ref{comparison}(b), we show that the single-path inference uses $3.8\times$ fewer parameters while maintaining the classification accuracy.

We compared our model with state-of-the-art sparse oblique trees such as TAO \cite{carreira2018alternating} and Bonsai \cite{kumar2017resource}, as well as other  ML models including logistic regression, kNN, multilayer perceptron (MLP), lightGBM \cite{ke2017lightgbm}, \textcolor{black}{FastGRNN \cite{kusupati2018fastgrnn}, SpArSe~\cite{fedorov2019sparse} and MODC \cite{gural2019memory}.} The settings of different models are as follows (also marked on Fig.~\ref{mnist_compare}): (1) \textbf{ResOT-$\ell_2$(n)}: the proposed resource-efficient oblique tree with $\ell_2$ regularization and a maximum depth of \textbf{n}. (2) \textbf{lightGBM(n)}: a boosted ensemble with \textbf{n} axis-aligned decision trees, where a single tree is a one-vs-all (ova) binary classifier. (3) \textbf{Bonsai}: the Bonsai model that is based on sparse-projected trees  \cite{kumar2017resource}. (4) \textbf{TAO(n)}: the alternating optimized oblique tree (TAO) with a maximum depth of \textbf{n}. TAO iteratively optimizes the decision functions of the internal nodes and uses the $\ell_1$ regularization to generate sparse connections. (5) \textbf{linear classifier}: one-vs-all logistic regression classifier. (6) \textbf{kNN(n)}: \textbf{n}-nearest neighbor classifier. (7) \textbf{MLP(n)}: multilayer perceptron with single hidden layer and \textbf{n} units. \textcolor{black}{(8) \textbf{SpArSe} and \textbf{MODC}: memory-optimized convolutional neural networks. (9) \textbf{FastGRNN}: gated recurrent neural network with optimized model size.} In all experiments, the weights were quantized to 4 bits (16 shared weights).  

With a shallow ResOT (maximum depth of 4), we achieved a 7.81\% test error with only 2.5kB of memory, which is better than  TAO \cite{carreira2018alternating} (10.2\%) at similar settings (maximum depth of 4). The test error can be further reduced by increasing the depth. We achieved a 4.94\% error with a depth of 7, which is better than  TAO  (5.69\%) with a deeper tree (depth of 12). We used the single-path inference scheme  to evaluate the test error of ResOT in these experiments. \textcolor{black}{
It should be noted that although deep learning methods based on CNN/RNN (e.g., SpArSe, MODC, FastGRNN) achieve a good performance with a small model size, they are likely to suffer from a higher computational complexity compared to DTs, since single-path inference is not possible in such networks and they typically require many multiplications and additions in their multi-layer architectures. 
On the contrary, the proposed ResOT  inherits the lightweight single-path inference thanks to its hierarchical structure, and outperforms state-of-the-art oblique tree-based models with low complexity such as Bonsai and TAO.}

\begin{figure}
  \centering
  \includegraphics[width=1\columnwidth]{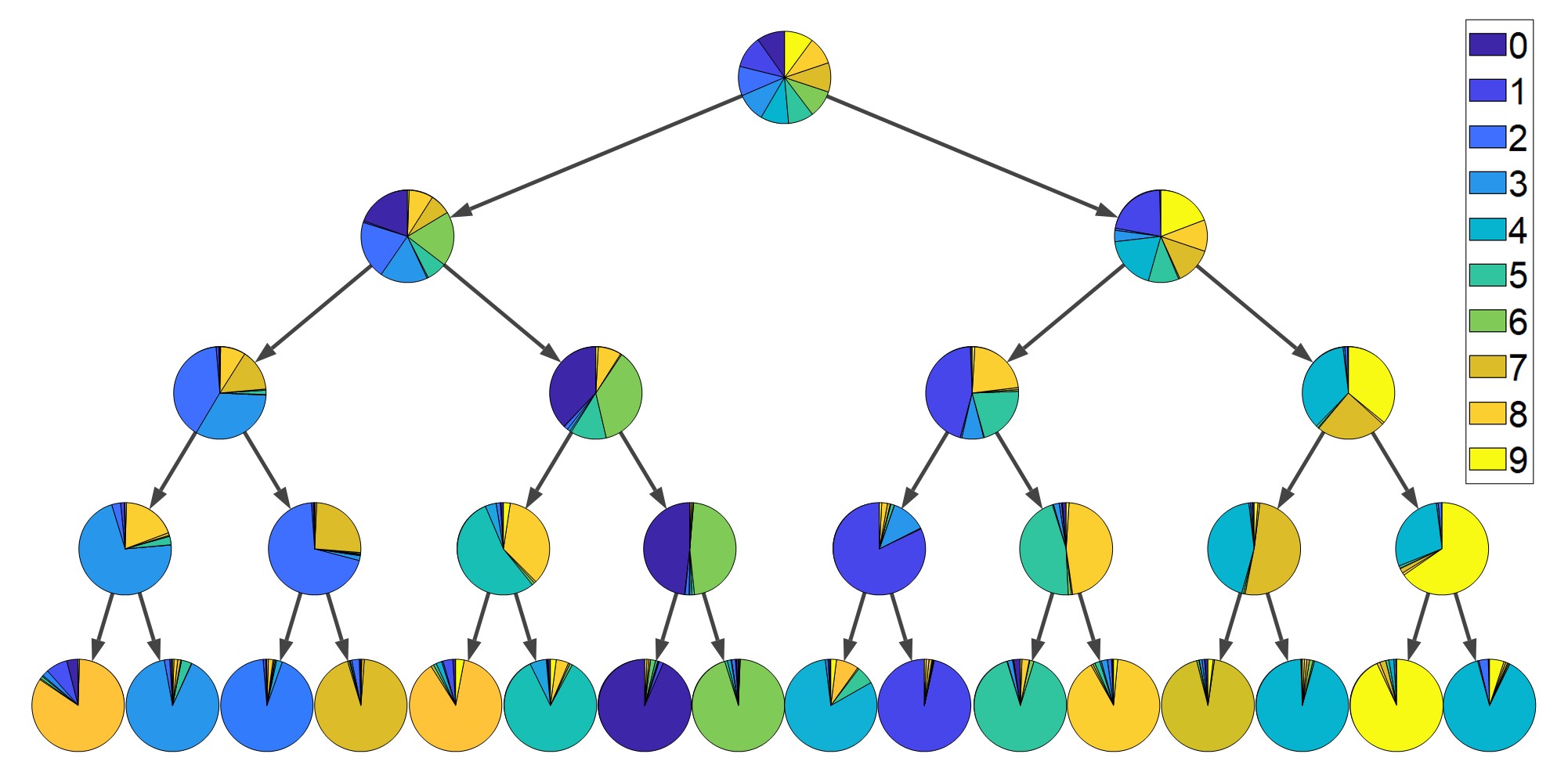}
  \caption{The visualization of ResOT trained on the MNIST dataset. Both internal and leaf nodes are represented by pie charts, indicating the class distribution of test samples. We can see that samples with different class labels are mixed in the root node. Following classification, however, each leaf node  has a dominant class label.}
  \label{vis}
  \vspace{-5mm}
\end{figure}

The decision process of ResOT is interpretable and can be easily visualized. Figure \ref{vis} shows the visualization of our oblique tree (max depth of 4) trained on the MNIST dataset. 
Both internal  and leaf nodes are represented by pie charts, showing the label distribution of the test data going through those nodes. The labels are from 0 to 9, indicating which digit the image is belonging to. 
By moving deeper in the tree, different digits tend to appear at different leaves. The leaf nodes are labeled by the dominant class passing through that node. Therefore, a single ResOT is capable of accurate multi-class separation with a shallow depth of 4.

\begin{figure*}[t]
    \centering
  \includegraphics[width=2\columnwidth]{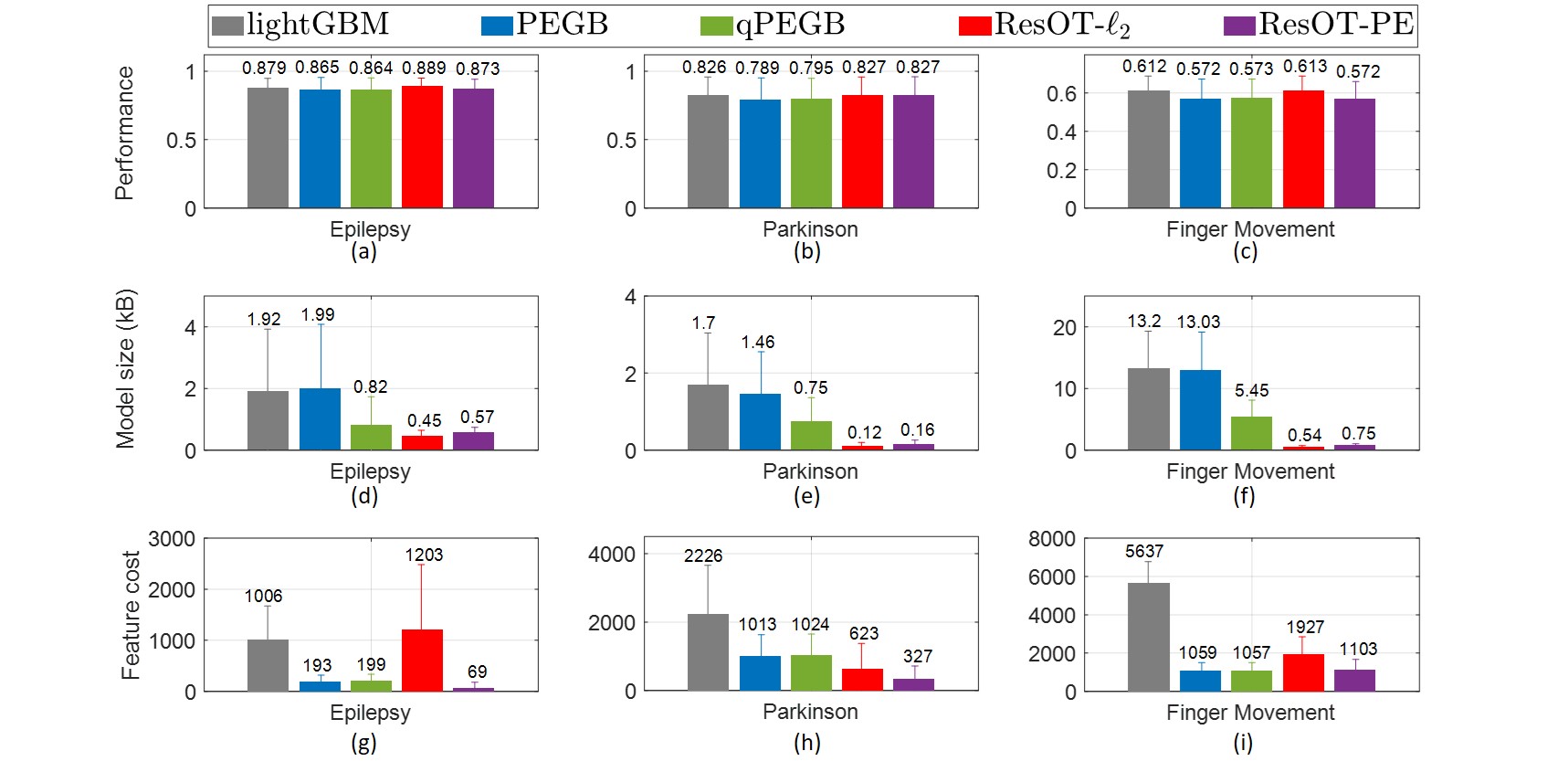} 
  \caption{The classification performance on neural  tasks. We implemented 5 different DT-based models including lightGBM \cite{ke2017lightgbm}, power-efficient gradient boosting (PEGB) \cite{zhu2019cost}, quantized PEGB (qPEGB) \cite{zhu2019cost}, and the proposed resource-efficient oblique tree with $\ell_2$ regularization (ResOT-$\ell_2$, this work) and with power-efficient regularization (ResOT-PE, this work). The models are compared in classification performance, model size, and feature cost (power), for each task. The error bars indicate the standard deviation across subjects and the average values are shown on the bars. The  size of memory and feature extraction cost are reduced with the proposed oblique tree-based models.}
  \label{main}
\end{figure*}

\vspace{-4mm}
\subsection{Performance on Neural Tasks}
In order to assess the  performance  of the proposed ResOT model in neural signal classification, we performed three experiments including epileptic seizure detection, Parkinsonian tremor detection, and finger movement classification. \textcolor{black}{
We used a block-wise approach to split the data into train and test sets. For seizure detection, the continuous iEEG recordings were divided into multiple blocks, where each block consisted of one seizure segment and the preceding non-seizure segment. For tremor and finger movement detection, the continuous LFP and ECoG recordings for each patient were divided into 5 equal-sized blocks, without shuffling. A 5-fold cross-validation method was used to report the classification performance for each task, where 80\% of blocks (rounded to the nearest integer) were used for training the model in each round and 20\% for testing, and the results of 5 rounds were averaged. For epilepsy patients with less than 5 seizures, the performance was evaluated based on a leave-one-out cross-validation (i.e., 2 blocks for training and 1 block for testing in a patient with 3 seizures).}
\textcolor{black}{The average training time of ResOT-$\ell_2$/ResOT-PE  
for seizure detection, tremor detection, and finger movement classification was 86.1/144.2, 4.2/5.2 and 36.4/41.6 seconds, respectively.}

Figure \ref{main} compares different tree-based machine learning models in terms of classification performance, model size, and feature extraction cost on three neural tasks. \textcolor{black}{We have previously shown that the ensemble of gradient boosted trees achieves a higher classification performance on these tasks (i.e., seizure detection \cite{shoaran2018energy}, tremor detection \cite{yao2020improved}, and finger movement classification \cite{yao2019enhanced}), compared to other ML models. Therefore, here we compare our approach against the lightGBM classifier \cite{ke2017lightgbm}, which is a fast and high-performance implementation of gradient-boosted trees}. The hyperparameters (maximum depth and tree count) for lightGBM were optimized in a subject-specific basis. Overall, we trained 4--20 trees for seizure detection, 5--40  for tremor detection, and 30--60 for finger movement classification, with maximum depths ranging from 2 to 6. The second model, Power-Efficient Gradient Boosting (PEGB), is a modified version of lightGBM that incorporates the feature  power dissipation in the objective function \cite{zhu2019cost}. In our previous work, we showed that this  approach can reduce the feature computational cost while maintaining the classification performance for both seizure and tremor detection tasks \cite{zhu2019cost}. However, even though PEGB can greatly reduce the hardware cost of a conventional gradient boosting model, its circuit implementation suffers from the large model size of the ensemble and limited scalability. To partially tackle this issue, quantized PEGB (qPEGB) with fixed-point arithmetic was proposed \cite{zhu2019cost} that improved the model size compared to PEGB. We quantized the thresholds and leaf weights with 12 bits for tremor detection, while 3/12 bits were used to quantize the leaf weights/thresholds for seizure detection and movement classification  \cite{zhu2019hardware}. 
Similarly, the maximum depth and number of trees for PEGB and qPEGB were optimized for each subject and both models were included in this study. 

Furthermore, we implemented the proposed ResOT model as described in Section \ref{Probabilistic}, with two different settings: (1) oblique tree with $\ell_2$ regularization (ResOT-$\ell_2$), and (2) oblique tree with a power-efficient regularization (ResOT-PE). \textcolor{black}{
Specifically, we regularized the oblique trees by tuning the weight pruning threshold and regularization coefficients ($\lambda$ in Eq. \ref{eq5}). We also limited the maximum tree-depth, while the final depth was determined by the algorithm under several constraints (pruning, regularization, maximum depth).
} Both ResOT-$\ell_2$ and ResOT-PE consist of a single oblique tree with a maximum depth of 4, while 4 bits were used to represent the shared weights. The regularization coefficients ($\lambda$ in Eq.~\ref{eq5}) and number of parameters (following weight pruning) were optimized for each patient. \textcolor{black}{Overall, the size of the weight matrix $\boldsymbol{\theta}$ (i.e., $ I \times D $) varies from 15$\times$565 to 15$\times$1409 for seizure detection, is 15$\times$37 for tremor detection, and varies from 15$\times$343 to 15$\times$577 for finger movement classification. After weight pruning, the weight matrix becomes sparse, with an average of 108.8/121.6, 57/89, and 270.2/455.1 non-zero elements  for ResOT-$\ell_2$/ResOT-PE on  three  neural tasks.}

We used the accuracy measure to evaluate the classification performance on finger movement task, Fig. \ref{main}(c), and F1 score for seizure, Fig. \ref{main}(a), and tremor detection tasks, Fig. \ref{main}(b), due to their highly imbalanced datasets. In our experiments, lightGBM achieved an average F1 score of 0.879($\pm$0.069) on seizure detection, 0.826($\pm$0.131) on tremor detection, and an average accuracy of 0.612($\pm$0.077) on finger movement classification task. However, rather than using an ensemble of boosted trees, we show that a comparable performance can be achieved using a single oblique tree with $\ell_2$ regularization (F1 scores of 0.889($\pm$0.059) for epilepsy and 0.827($\pm$0.131) for PD, and an accuracy of 0.613($\pm$0.076) for finger movement). 
\textcolor{black}{Furthermore, the sensitivity and specificity of these models in seizure and tremor detection are reported in Table.~\ref{sen_spe}.}
With power-efficient approach (PEGB, qPEGB, ResOT-PE), we made a trade-off between classification performance and hardware cost. 
Overall, the proposed ResOT-PE model achieves a comparable classification performance as lightGBM on all tasks, while significantly reducing the model size and power consumption, as later discussed in this Section.  

Figures \ref{main}(d-f) compare the model size for different ML algorithms. As shown in these figures, ResOT-$\ell_2$ is the most memory-efficient model, with 0.45($\pm$0.19), 0.12($\pm$0.08), and 0.54($\pm$0.19)~kB model sizes for the three performed tasks, respectively. Compared with lightGBM, ResOT-$\ell_2$ reduces the model size \textcolor{black}{by 4.3$\times$ for seizure detection, 14.2$\times$ for tremor detection, and 24.4$\times$ for finger movement classification}. The ResOT-PE has a slightly larger model size compared to ResOT-$\ell_2$ \textcolor{black}{(memory saving of 3.4$\times$, 10.6$\times$, and 17.6$\times$ for seizure detection, tremor detection, and finger movement classification, compared to lightGBM)}, since the power consumption is optimized along with memory. The model size of PEGB is comparable to lightGBM, while a reduction factor of 2.3$\times$ can be achieved by quantizing the parameters (qPEGB). 

As discussed in Section \ref{hce} and IV-A, the power consumption for feature extraction was minimized through our power-efficient regularization framework. Figures \ref{main}(g-i) depict the inference power for various models on the three neural tasks. The power during inference was calculated based on a single-path scheme for all models, by summing up the power consumption of features visited along the decision path. For example, for the case of ResOT, the power cost for a test sample is calculated by summing up the cost of extracted features:  $\sum_{i=1}^{I} \sum_{j=1}^{D} \beta_j C_{i,j}$, where $C_{i,j}=$ 0 or 1 indicates whether the internal node $i$ is on the decision path and  feature $j$ is extracted in that node.
Compared to the benchmark model (lightGBM), PEGB achieves an average power reduction of 4.2$\times$ on these tasks. Compared to lightGBM, the ResOT-PE reduces the estimated power by \textcolor{black}{ 14.6$\times$ for seizure detection, 6.8$\times$ for tremor detection, and 5.1$\times$ for finger movement classification}. The ResOT-PE model obtains the lowest feature extraction cost on average, while offering a considerably smaller model size compared to qPEGB. 

\begin{table}[t!]
\textcolor{black}{
\caption{Sensitivity and Specificity for Seizure and Tremor Detection}
\vspace{-3mm}
\begin{center}
\scalebox{0.90}{
\begin{tabular}{c|c|c|c|c}
\hline \hline
\multirow{2}{*}{Models}& \multicolumn{2}{c|}{Seizure Detection} & \multicolumn{2}{c}{Tremor Detection} \\ \cline{2-5}
& Sensitivity  & Specificity & Sensitivity & Specificity\\
\hline
lightGBM &  0.881$\pm$0.069 &0.984$\pm$0.016  & 0.890$\pm$0.103 & 0.281$\pm$0.263 \\
PEGB & 0.881$\pm$0.077&  0.974$\pm$0.036& 0.838$\pm$0.171 & 0.274$\pm$0.259   \\
qPEGB & 0.881$\pm$0.077 & 0.974$\pm$0.036 &  0.850$\pm$0.157 & 0.262$\pm$0.252 \\
\textbf{ResOT-$\ell_2$} & 0.877$\pm$0.076 &0.986$\pm$0.012  &  0.897$\pm$0.097 & 0.274$\pm$0.252  \\
\textbf{ResOT-PE} & 0.860$\pm$0.084 & 0.989$\pm$0.012 &0.901$\pm$0.091 & 0.247$\pm$0.260  \\
\hline \hline 
\end{tabular}
}
\label{sen_spe} \vspace{-5mm}
\end{center} }
\end{table}

 \vspace{-3mm}
\subsection{Power-Efficient Inference}
\begin{figure}
    \centering
  \includegraphics[width=0.99\columnwidth]{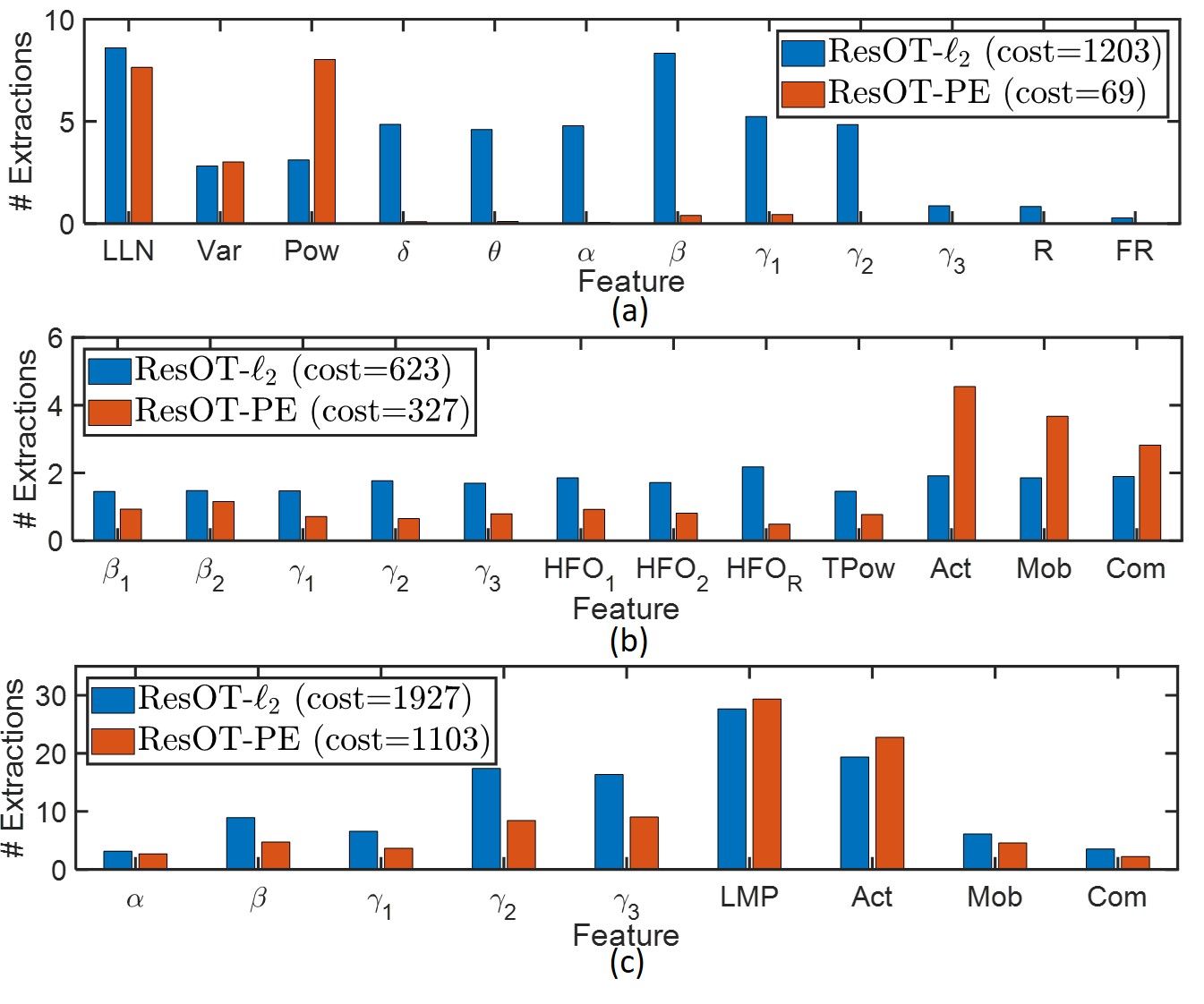}  
  \vspace{-8mm}
  \caption{Illustration of the proposed power-efficient framework in terms of feature extraction count, by comparing the $\ell_2$ and power-efficient regularization schemes. The bars indicate the total number of extractions for each feature along the decision path, that is averaged over subjects, for (a) seizure detection, (b) tremor detection, and (c) finger movement classification tasks.} 
  \label{hist}
 \vspace{-3mm}
\end{figure}

To better demonstrate the superior power efficiency of  ResOT-PE  over ResOT-$\ell_2$, we took a closer look into the structure of these models and the extracted features in the oblique nodes. Figure \ref{hist} shows the average number of extractions for each feature, using the $\ell_2$ and power-efficient regularization schemes. By adopting the power-efficient regularization term (Eq. \ref{eq16}) with the weight pruning framework, the power-hungry features are assigned smaller weights, and as a result, they are less likely to survive during the weight pruning process. 
As depicted in Table. \ref{tab2}, each feature is associated with a normalized power cost, including its static and dynamic power. Taking the seizure detection task as an example, power, variance, and line-length have a low complexity, while band power features consume the highest energy due to multiplications and additions required for FIR filtering. In the power-efficient learning scheme, hardware-friendly features are favored over  costly ones, and the model tends to use a higher number of these features compared to band powers. Similarly, we see a smaller number of band power extraction for tremor detection and finger movement classification, while a higher number of Hjorth activity, mobility, and LMP features are extracted in these tasks.
Compared with ResOT-$\ell_2$, the ResOT-PE model reduced the power cost by \textcolor{black}{ 17.4$\times$, 1.9$\times$, and 1.7$\times$ }for these neural tasks, respectively, as shown in Fig. \ref{hist}. 

It should be noted that in this work, we implemented a specialized digital hardware to measure the power of each feature as a stand-alone block, including its dynamic and static power. However, in a full system implementation, various resource optimization and power reduction techniques (e.g., mixed-signal design,  re-using the common blocks between different features, power and clock gating) can be employed to further reduce the total energy of the system. 

 \vspace{-3mm}
\section{Discussion}
The proposed ResOT model could be considered as a combination of decision trees and neural networks, that is further optimized for low-power implementation. With the mini-batch gradient descent training, ResOT is also compatible with an  online learning framework  where the model could dynamically adapt to the non-stationary nature of neural signal, which is the focus of our future work. 
In this Section, we discuss the contributions of this paper and the benefits of ResOT in terms of memory and power efficiency, as well as the hardware overhead for implementing oblique splits, with a focus on the above neural  classification tasks. 
 \vspace{-3mm}
\subsection{Hardware Improvements}
The benefits of the proposed ResOT model for hardware implementation are as follows: \textbf{1. Memory Efficiency;} Compared with conventional oblique trees, the proposed ResOT model is compressed via weight pruning and sharing. The resulting parameter matrix is sparse and can be efficiently stored in an on-chip memory such as SRAM. Compared to our earlier work that integrated 8 gradient-boosted trees with 1kB of memory for epilepsy \cite{shoaran2018energy}, this work requires \textcolor{black}{2.2$\times$ less memory on average (ResOT-$\ell_2$)}, for storing the parameters of a single oblique tree. \textbf{2. Lightweight Inference;} Compared to a neural network (that could be compressed with similar techniques), our model benefits from the hierarchical structure of trees. During inference, we only need to follow a single root-to-leaf path to make predictions, without visiting the rest of the model. This lightweight inference can potentially improve the energy efficiency for implantable and edge applications. \textbf{3. Cost-Aware Learning;} With cost-aware learning, ResOT learns to prioritize the lower-cost features during inference. We used the power consumption as a measure of cost in this work, and ResOT-PE achieved one of the lowest power consumptions among the analyzed models on all target tasks.
\textcolor{black}{The pros and cons of different tree-based models are summarized in Table.~\ref{comp_model}.} 

\begin{table}[h]
\textcolor{black}{
\caption{Comparison of Different Tree-based Models}
\vspace{-3mm}
\begin{center}
\scalebox{0.85}{
\begin{tabular}{c|c|c|c|c}
\hline \hline
Model & Cost-aware & Model Compression & $\#$ of Trees & Split Type   \\
\hline
lightGBM & \xmark    & \xmark  & $>1$  & Axis-aligned\\
PEGB & \cmark & \xmark   & $>1$ & Axis-aligned\\
qPEGB & \cmark & \cmark $^\dagger$  &  $>1$&  Axis-aligned\\
\textbf{ResOT-$\ell_2$} & \xmark & \cmark $^\ddagger$   & 1& Oblique\\
\textbf{ResOT-PE} & \cmark & \cmark $^\ddagger$ & 1& Oblique\\
\hline \hline 
\multicolumn{5}{l}{$^\dagger$ weight and threshold quantization} \\
\multicolumn{5}{l}{$^\ddagger$ weight pruning and sharing}
\end{tabular}
}
\label{comp_model} \vspace{-5mm}
\end{center} }
\end{table}


\vspace{-3mm}
\subsection{Benefits of Oblique Splits}
Conventional DT-based models such as random forest and lightGBM are composed of axis-aligned trees with one feature-threshold pair in each internal node. The decision boundaries of axis-aligned trees are parallel to the axes of the feature space. As a result, such topologies 
ignore the correlations between features and might be suboptimal for classifying highly correlated data. To tackle this issue, oblique trees use a linear model as the split function. Thus, multiple features are combined at the internal node and the decision boundary is an oblique hyperplane that can better adapt to the various distributions of input data. Therefore, oblique trees perform better on signals with strongly correlated features~\cite{menze2011oblique}. This is typically the case for neural signal classification tasks. For example, line-length and variance both describe the signal variations and show a positive correlation during a seizure, although their mathematical definitions are different.
Moreover, the correlation between neural signals recorded by adjacent channels of an electrode array (depending on the spacing of electrodes), leads to a correlation between  corresponding features. Thus, oblique trees proposed here are favored over axis-aligned trees for neural signal classification tasks (e.g., ResOT-$\ell_2$/ResOT-PE perform slightly better than lightGBM/PEGB in terms of F1 score, as shown in Fig.~\ref{main}(a-c)).

\begin{table}[t!]
\caption{The Overhead Cost of Oblique Nodes in ResOT}
\vspace{-3mm}
\begin{center}
\scalebox{0.93}{
\begin{tabular}{c|c|c|c|c}
\hline \hline
\multirow{2}{*}{Task}& \multicolumn{2}{c|}{ResOT-$\ell_2$} & \multicolumn{2}{c}{ResOT-PE} \\ \cline{2-5}
& \# Mult.  & \# Add. & \# Mult. & \# Add.\\
\hline
Epilepsy & 108.8 & 102.6 & 121.6& 107.1  \\
Parkinson & 57 & 46.8 & 89 & 75.5  \\
Finger Movement & 270.2 & 260 & 455.1 & 440.1  \\
\hline \hline 
\end{tabular}
}
\label{overhead} \vspace{-5mm}
\end{center}
\end{table}

Indeed, axis-aligned trees can be considered as a subset of oblique trees and have the advantage of fast training and easy interpretation. 
In this paper, we show that ResOT is a memory- and power-efficient alternative for large axis-aligned ensembles, and is particularly useful for neural classification tasks that require many trees.

\vspace{-3mm}
\subsection{Hardware Complexity of Oblique Nodes}
Axis-aligned decision trees use simple comparators  at their internal nodes to compare a feature with a threshold. 
Alternatively in this work, we used oblique trees trained with a probabilistic routing. While a  \textit{sigmoid} function needs to be calculated in a probabilistic training phase,  it is simplified to a comparison during single-path inference (i.e., $\boldsymbol{x_n^{\top}\theta_i}>0$ or $\leq0$ ?), since the test samples only travel through the most probable path. Thus, similar to gradient-boosted trees \cite{shoaran2018energy}, the hardware complexity  is dominated by feature extraction process and comparators can be ignored in total power estimation.

However, oblique trees still require a weighted sum of features as input to the comparator. These features are associated with the non-zero elements \textcolor{black}{of the sparse weight matrix $\boldsymbol{\theta}$} following weight pruning. In addition to the hardware cost for features (Fig.~\ref{main}), the overhead cost of implementing the weighted sum should be considered for an oblique tree. \textcolor{black}{Table.~\ref{overhead} summarizes the average overhead cost of ResOT-$\ell_2$ and ResOT-PE on three neural tasks, by calculating  the total number of  multiplications and additions required to generate the weighted sum of features in an oblique tree (i.e., calculated for 15 nodes in a tree of depth 4). Since each non-zero element is associated with a multiplication, the number of additional multiplications is inversely proportional to the sparsity of the weight matrix $\boldsymbol{\theta}$.}
Interestingly, it can be shown that the hardware cost for oblique node implementation is negligible compared to a single feature extraction. Here, we consider the case of neural data classification using oblique trees, where the input signal is sampled at $f$ Hz with a window size of $W$ used for feature extraction. To implement a band power feature (as a comparison), the signal is first passed through a digital FIR filter. A total number of $t \times N$ multipliers and $(t-1) \times N$ adders are required to filter the neural signal, where $t=30$ represents the number of FIR taps and $N=f\times W$ indicates the number of samples in a window of signal. We then calculate the power of the filtered signal, which requires an additional $N$ multipliers and $N-1$ adders. In total, the extraction of a band power feature requires $(t+1) \times N$ multiplications and $t \times N -1$ additions. Assuming the worst case  overhead cost for ResOT-PE applied to finger movement classification (Table.~\ref{overhead}) with $f=1$kHz and $W=0.2s$, the extraction of a single band power feature requires 6200 multiplications and 5999 additions, whereas the ResOT-PE evaluation only requires 455.1 multiplications and 440.1 additions on average (over subjects) to linearly combine the features in the oblique nodes. Thus, the overhead cost is  lower than feature cost of a single band power feature by over an order of magnitude. \textcolor{black}{Moreover, in the single-path inference scheme, a maximum of 4 nodes are sequentially processed per tree, which could further reduce the complexity of the classifier during inference. }
Therefore, the overhead cost of oblique nodes in this single-tree scheme is marginal and will not burden the resource-efficient implementation of proposed OT model.


\vspace{-2mm}
\section{Conclusion}
In this paper, we proposed the ResOT model, a hardware- and memory-efficient approach for neural signal classification. Weight pruning and sharing were applied together with power-efficient regularization to compress the tree and enable cost-aware learning. Being trained with a probabilistic routing,  ResOT  benefits from a  single-path inference scheme, enabling its lightweight implementation.  Testing on three neural signal classification tasks with 31 patients, our model outperformed the state-of-the-art ensemble of boosted trees in both model size and power consumption. Resource-constrained applications such as neural implants and IoT devices could significantly benefit from the proposed hardware-friendly classifier.

\vspace{-1mm}
\section*{Acknowledgment}
This work was supported by a Google faculty research award in machine learning.

\ifCLASSOPTIONcaptionsoff
  \newpage
\fi



%
\bibliographystyle{IEEEtran}
\bibliography{cit}


\end{document}